\documentclass[10pt,twocolumn,letterpaper]{article}

\usepackage{iccv}
\usepackage{times}
\usepackage{epsfig}
\usepackage{graphicx}
\usepackage{amsmath}
\usepackage{amssymb}

\usepackage[switch]{lineno} 
\usepackage{subcaption}
\usepackage{amsmath}
\usepackage{amsfonts}
\usepackage{algorithmicx,algorithm}
\usepackage[noend]{algpseudocode}
\usepackage{booktabs}
\usepackage{multirow}
\usepackage{makecell}
\usepackage[inline]{enumitem}
\newcommand{\discriminator}{\mathcal{D}_\omega}
\newcommand{\transition}{\mathcal{T}_\mu}
\newcommand{\policy}{\pi_\theta}
\newcommand{\posterior}{q_\varphi}
\algnewcommand{\LineComment}[1]{\State \(\triangleright\) #1}
\let\sss= \scriptscriptstyle
\graphicspath{{fig/}}
\usepackage[dvipsnames]{xcolor}

\usepackage[pagebackref=true,breaklinks=true,letterpaper=true,colorlinks,bookmarks=false]{hyperref}

\iccvfinalcopy 

\def\iccvPaperID{5878} 

\ificcvfinal\fi

\begin{document}

\title{Procedure Planning in Instructional Videos \\ via Contextual Modeling and Model-based Policy Learning}

\author{\quad Jing Bi \;\;\quad Jiebo Luo  \quad Chenliang Xu\\
University of Rochester\\
{\tt\small jing.bi@rochester.edu} \qquad {\tt\small jiel@cs.rochester.edu}\qquad {\tt\small chenliang.xu@rochester.edu}
}

\maketitle
\ificcvfinal\thispagestyle{empty}\fi

\begin{abstract}
	Learning new skills by observing humans' behaviors is an essential capability of AI.
	In this work, we leverage instructional videos to study humans' decision-making processes, focusing on learning a model to plan goal-directed actions in real-life videos.
	In contrast to conventional action recognition, goal-directed actions are based on expectations of their outcomes requiring causal knowledge of potential consequences of actions.
	Thus, integrating the environment structure with goals is critical for solving this task.
	Previous works learn a single world model will fail to distinguish various tasks, resulting in an ambiguous latent space; planning through it will gradually neglect the desired outcomes since the global information of the future goal degrades quickly as the procedure evolves.
	We address these limitations with a new formulation of procedure planning and propose novel algorithms to model human behaviors through Bayesian Inference and model-based Imitation Learning.
	Experiments conducted on real-world instructional videos show that our method can achieve state-of-the-art performance in reaching the indicated goals.
	Furthermore, the learned contextual information presents interesting features for planning in a latent space.
\end{abstract}
\vspace{-4mm}
\section{Introduction}
\label{sec:intro}
\begin{figure}[t]
	\includegraphics[width=\columnwidth]{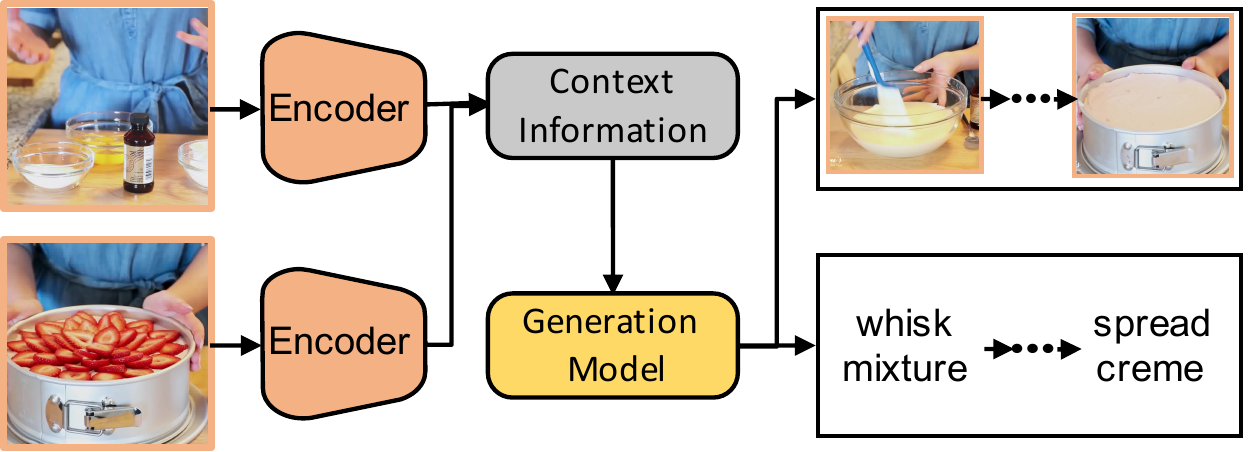}
	\caption{\textbf{Overview of our proposed method.} Given a starting observation (top-left image) and a desired visual goal (bottom-left image), we extract the contextual information of the planning trajectory upon which the Generation Model outputs a sequence of actions. The model is responsible for learning plannable latent representations with a focus on procedures and action consequences. Thus, we can retrieve images of intermediate steps (top-right images).}
	\label{fig:overview}
	\vspace{-6mm}
\end{figure}

Humans can learn new skills by watching demo videos. Although this seems natural to human, it is challenging for AI.
We have seen rich works on modeling human behaviors from videos with the majority focusing on recognizing actions~\cite{kong2018human,huang2019learning,martinez2019action}.
However, solely perceiving what actions are performed without modeling the underlying decision-making process is insufficient for AI to learn new skills.
The next-generation AI needs to figure out what actions are necessary to achieve the desired goals~\cite{casas2018intentnet} with the consideration of actions' potential consequences.
In this paper, we focus on learning the goal-directed actions from instructional videos.
Recently, Chang~\etal~\cite{chang2019procedure} proposed a new problem known as procedure planning in instructional videos.
It requires a model to 1) plan a sequence of verb-argument actions and 2) retrieve the intermediate steps for achieving a given visual goal in real-life tasks such as making a strawberry cake (see Fig.~\ref{fig:overview}).
This task is different from the typical image-language translation problem in the way that certain actions can be exchanged to achieve the same goal (\eg, the order of \textit{adding salt} and \textit{sugar} usually does not matter), making it difficult to predict the same action sequence as ground-truth using sequence mapping.

Moreover, sequence-to-sequence based structure, suitable for modeling events that tend to occur in sequence with high probability, is thought to involve no consideration of the likely outcome~\cite{RAAB2018279}.
Therefore, we formalize this task as a planning problem with focus on two different sequential patterns that can be easily observed in Fig.~\ref{fig:stat}:
In the context of \textit{making a cake}, \textit{mixing ingredients} and \textit{washing cherries} are interchangeable, \ie, short-term action separation, but both should be ahead of the action \textit{putting cherries on the top}, \ie, long-term action association.
\begin{figure*}[ht]
	\includegraphics[width=\textwidth]{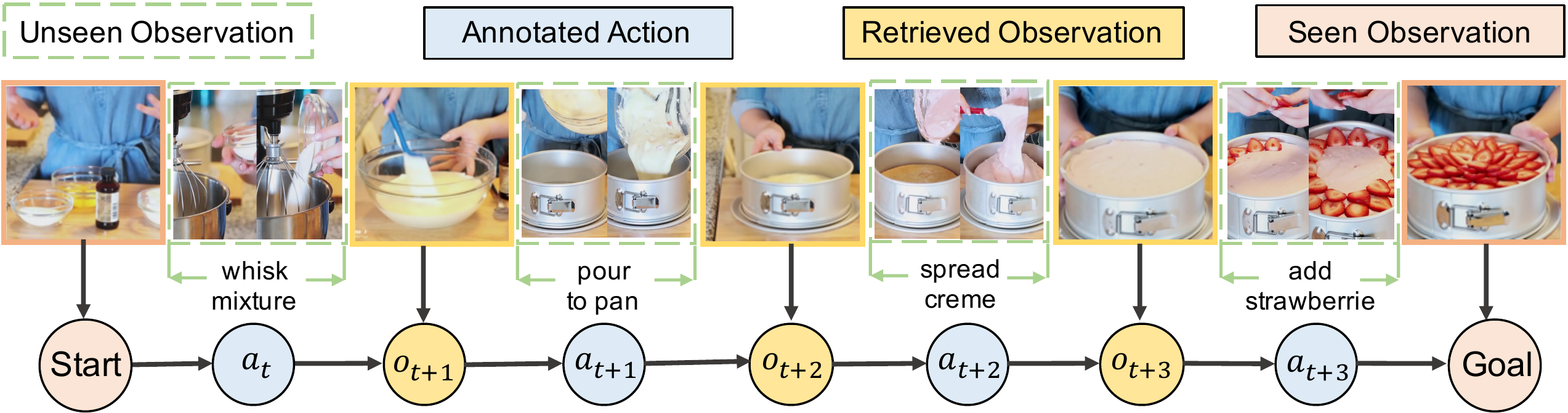}
	\caption{\textbf{Procedure planning example.} Given a starting observation (picture of food ingredients) and a visual goal (picture of a made cake), the model needs to learn how to complete real-world tasks such as making a cake by planning a sequence of actions  $a_{\sss{1:T}}$ (blue circles) and retrieving the intermediate observations $o_{\sss{2:T-1}}$ (yellow circles).}
	\label{fig:showcase}
	\vspace{-6mm}
\end{figure*}

Inspired by Raab \etal~\cite{RAAB2018279}, we think that when performing goal-directed tasks, it is beneficial to consider both the task contextual information and the potential action consequences. 
Contextual information here refers to the time-invariant knowledge (not changed during planning) that distinguishes a particular task from the others.
For example, if we know the goal is to \textit{make a cake} as shown in Fig.~\ref{fig:overview}, it is less likely to plan an action like \textit{putting it on the grill}.

Therefore, we model the dependency between the actions and different goals as the long-term action association in a Bayesian framework. 
As we show later in the experiments, this serves a few purposes:
\begin{enumerate*}[label=\alph*)]
	\item it provides a more structured representation for the subsequent policy learning; 
	\item we can sample from the posterior distribution for more diverse trajectories to facilitate the action exploration; and
	\item compared with the noisy pixel space, feature distances in the learned latent space are more meaningful.
\end{enumerate*}
To achieve short-term action separation, we model the action sequence as a Markov Decision Process (MDP) as shown in Fig.~\ref{fig:showcase}, where the future action depends only upon the present state.
Besides, because goal-directed actions are often selected based on expectations of their consequent outcomes~\cite{daw2005uncertainty}, we propose to incorporate a transition model into the Imitation Learning (IL) framework~\cite{NIPS2016_6391,kaiser2019model} so that we can explicitly model the environment jointly with policy learning. 

This approach brings following advantages: 
\begin{enumerate*}[label=\alph*)]
	\item it helps policy to ﬂexibly pursue a goal by leveraging causal knowledge of the actions' potential consequences;
	\item when model-based simulations produce states with alternative actions, the discrimination and selection between actions allow an agent to find the currently most desired outcome~\cite{liljeholm2018instrumental,sutton1991dyna}; and 
    \item it bypasses the need of an interactive environment that is required by classic planning algorithms~\cite{silver2017mastering,rhinehart2017first}, making it suitable for modelling the web videos.
\end{enumerate*}

We demonstrate the effectiveness of our approach by evaluating it on a real-world instructional video dataset~\cite{crosstask} (an example is shown in Fig.~\ref{fig:showcase}).
The results on the procedure planning task show that our learned model can uncover the underlying human decision-making processes. Furthermore, the results on the challenging walk-through planning task~\cite{kurutach2018learning} confirm that our model learns meaningful representations of the environment dynamics, which is crucial for efficient plannings in the latent space.
Finally, the visualization of contextual information indicates that our proposed encoder structure can learn a concise representation to capture distinct knowledge of different real-world tasks.
The main contributions of our work are summarized as follows:
\begin{enumerate*}[label=\alph*)]
	\item we propose a novel method to address the procedure planning problem, which combines Bayesian Inference with Model-based Imitation Learning; 
	\item we propose a neural network structure based on variational inference that learns to embed sufficient information to convey the desired task, incorporating the visual observations' uncertainty; and 
	\item we propose two model-based IL algorithms that explicitly learn the environment dynamics (in either a stochastic or deterministic way) and integrate with the transition model to simultaneously learn a plannable latent representation for accurate planning.
\end{enumerate*}

\section{Related Work}
\label{sec:related}
\noindent \textbf{Vision-based Human Behavior Understanding.}
Our planning tasks are highly related to a popular AI research area: building a machine that can accurately understand humans' actions and intentions.
The intention can be seen as the sequence of actions needed to be taken to achieve an objective~\cite{casas2018intentnet}. 
To understand human attention, Zhang \etal~\cite{Zhang_2017_CVPR} proposed the Deep Future Gaze model to predict the gaze location in multiple future frames conditioned on the current frame. Forthermore, Wei \etal~\cite{wei2018and} utilized a hierarchical graph that jointly models attention of the gaze and intention of the performing task from a RGB-D video.
Rhinehart \etal~\cite{rhinehart2017first} proposed an online inverse reinforcement learning method to discover rewards for modeling and forecasting first-person camera wearer's long-term goals, together with locations and transitions from streaming data.
Merel \etal~\cite{merel2017learning} extended the Generative Adversarial Imitation Learning (GAIL)~\cite{NIPS2016_6391} framework to learn human-like movement patterns from demonstrations consisting of only partial observations.
Unlike these previous works that predict the future, we try to understand human behaviors by learning their goal-direction actions. 

\noindent \textbf{Deep Reinforcement Learning.}
Reinforcement learning (RL) is often employed to learn and infer the MDP model simultaneously, which is a natural way to understand how humans learn to optimize their behaviors in an environment~\cite{sutton2018reinforcement}.
Recently, combined with deep learning, DRL is leveraged to solve several vision problems such as Visual Tracking~\cite{supancic2017tracking}, Video Summarization~\cite{zhou2018deep}, Stroke-based rendering~\cite{huang2019learning}, and Vision-based Navigation~\cite{nguyen2019vision}.
For semantic-level video understanding, DRL can also play an important role.
For instance, it is utilized for Activity Localization~\cite{wang2019language}, Natural Language Grounding~\cite{he2019read}, and Video Description~\cite{wang2018video}.
However, these works often require expert knowledge to design a useful reward function, whose goal is to learn a behavior that maximizes the expected reward.
In contrast, we work on the IL problem without explicit usage of a hand-craft reward.
Our work is most closely related to inverse RL~\cite{rhinehart2017first,abbeel2004apprenticeship} and contextual RL\cite{yu2019meta}.
However, the major difference is that we focus on learning from collected dataset which is crucial for applications when online interaction is not permitted, \eg, safety-critical situation.

\noindent \textbf{Planning in Latent Space.}
Planning is a natural and powerful approach to decision-making with known dynamics, such as game playing and simulated robot control.
To plan in unknown environments, the agent needs to learn the environment dynamics from previous experiences.
Recent model-based RL schemes have shown promise that deep networks can learn a transition model directly from low-dimensional observations and plan with the learned model~\cite{zeng2017visual,chang2019procedure,hafner2019learning}.
A closely related method is Universal Planning Networks (UPN)~\cite{srinivas2018universal} that learns a plannable latent space  with gradient descent by minimizing an imitation loss, i.e., learned from an expert planner.
Plannable means the learned representations are structured to perform a classic planning algorithm~\cite{kurutach2018learning}. 
Our method further incorporate the contextual knowledge of assigned task to the latent space and remove the assumption of differentiable action space.
Another line of work is causal InfoGAN~\cite{kurutach2018learning}, which tries to capture the relations between two sequential images and models the causality of the simulation environment in an unsupervised learning manner.
Similarly, our Ext-MGAIL model also focus on the stochastic transition model. 
However, making predictions in raw sensory space is unnecessarily hard~\cite{finn2016unsupervised}, we predict low dimensional latent representations for future state and plan upon it.
\section{Methods}
\label{sec:method}
We consider a set-up similar to Chang \etal~\cite{chang2019procedure}: we have access to $K$ trajectories $\{(o_{\sss{1:T}}^j,a_{\sss{1:T}}^j)\}_{j=0}^{\sss K}{\sim}\pi_{E}$  collected by an expert trying to achieve different tasks.
Given a starting visual observation $o_1$ and a visual goal $o_{\sss T}$ that indicates for a particular task, we want to learn a plannable representation upon which goal-directed actions are planned to perform two complex planning tasks (Fig.~\ref{fig:showcase}):
\begin{enumerate*}[label=\alph*)]
	\item procedure planning: generate a valid sequence of actions $a_{\sss{1:T}}$ to achieve the indicated goal; and 
	\item walk-through planning: retrieve the intermediate observations $o_{\sss{2:T-1}}$ between the starting $o_{\sss{1}}$ and the goal.
\end{enumerate*}
Our key insight is that by decomposing the procedure planning problem in Eq.~\ref{equ:formulation} into two sub-problems, we can decouple representation learning into two parts: 
\begin{enumerate*}[label=\alph*)]
	\item inferring the time-invariant contextual information that conveys the task to achieve; and 
	\item learning the time-varying plannable representations related to the decision-making process and environment dynamics.
\end{enumerate*}
In this way, both representations can be further used to retrieve $o_{\sss{2:T-1}}$ for solving the walk-through planning.

\begin{figure*}[ht]
	\centering
	\includegraphics[width=0.75\textwidth]{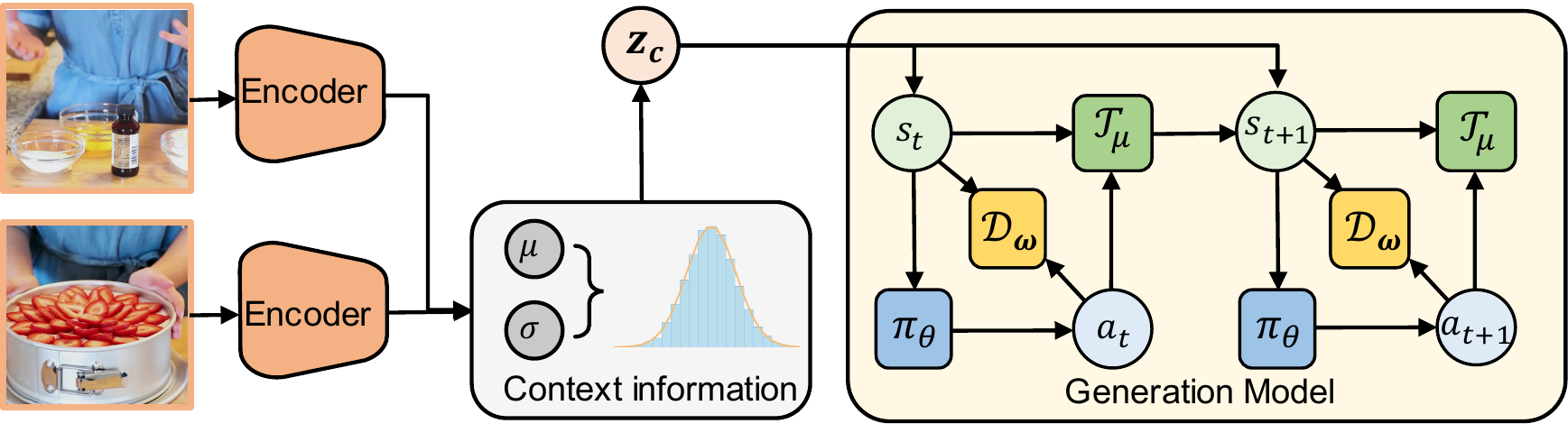}
	\caption{\textbf{The overall architecture:} given the initial and the goal observations, two parallel encoders will parameterize the mean and log-variance of the Gaussian distribution. The context variable will then be sampled from this distribution and fed into the generation model to  roll out a trajectory. We use a discriminator that tries to distinguish the state-action pairs from the expert or the learned policy, which serves as the local reward function. }
	\label{fig:model}
	\vspace{-3mm}
\end{figure*}

As shown in the overall architecture Fig.~\ref{fig:model}, we assume that the contextual information contains all the details an agent need for achieving the desired goal.
Hence, we formulate the procedure planning problem $p(a_{\sss{1:T}}|o_{\sss{1}},o_{\sss{T}})$ as:
\begin{equation}
	\label{equ:formulation}
	p(a_{\sss{1:T}}|o_{\sss{1}},o_{\sss{T}})=\iint p(a_{\sss{1:T}},s_{\sss{1:T}}|z_c)p(z_c|o_{\sss 1},o_{\sss T})\mathrm{d}s_{\sss{1:T}} \mathrm{d}z_c ,
\end{equation}
where we donate $z_c$ as the context variable that conveys the desired task, $p(z_c|o_1,o_T)$ as the inference model for modeling posterior distribution over the context variable and given observations and $p(a_{\sss{1:T}},s_{\sss{1:T}}|z_c)$ as the generation model that plans a sequence of actions and hidden states that transfer the initial state to the desired outcome.
In the following sections, we will first discuss how to infer the contextual information.
We will then solve the second sub-problem by imitating human behaviors with consideration of offline policy evaluation~\cite{levine2020offline}, and utilize Hindsight Experience Replay (HER)~\cite{andrychowicz2017hindsight} to better leverage the expert demonstrations.
Lastly, we will discuss how to solve the walk-through planning problem with learned model.

\subsection{Inference Model}
\label{subsec:infer}
As visualized in Fig.~\ref{fig:model} the action $a_t$ at time-step $t$ is solely governed by the current state $s_t$ that contains the information of current observation and the information regarding the desired goal.
We want $z_c$ to represent the contextual information for achieving the goal, which should be time-invariant, and the hidden state $s_t$ to contain the time-varying information for the decision-making process.
To achieve this separation, the hidden states are only allowed to condition on $z_c$; thus, all the information about the goal must pass through $z_c$ to avoid a shortcut from the observations to actions.
In this way, $s_t$ will be the only time-dependent hidden variable used to recover the actions, and we reserve $z_c$ for compressing everything else.
However, the true posterior distribution $p_\phi(z_c|o_1,o_{\sss T})$ from video frames is analytically intractable; thus, we use variational inference to approximate posterior distribution from given observations.
Note here we use the raw pixel observations $o_1$ and $o_{\sss T}$ for planning, which is different from compared methods which use pre-computed visual features.
The model involves an encoder (shown in Fig.~\ref{fig:model}) that models the approximation distribution $q_\varphi(z_c|o_1,o_t)$ and a decoder (omitted from Fig.~\ref{fig:model} for simplicity) that models the prior $ p_\phi(o_1,o_t|z_c)$. 
It can be seen as a two-head Variational Auto-Encoder (VAE)~\cite{kingma2013auto} with one head encoding $o_1$ and the other for $o_T$, and we call it a predictive VAE.
We jointly optimize $\phi$ and $\varphi$ by maximizing the evidence lower bound:
\vspace{-4mm}
\begin{equation}
  	\label{equ:decompose}
	\resizebox{0.90\hsize}{!}{
		$l(\phi,\varphi)=\mathbb{E}_{q_\varphi}[\log p_\phi(o_1,o_{\sss T}|z_c) ]-{\rm KL}(q_\varphi(z_c|o_1,o_{\sss T})||p(z_c)) $\;,
	}
\end{equation}
where we assume $p(z_c)$ is a Gaussian prior parameterized by the context variable. 
By training in this way, the encoder $q_\varphi(z_c|o_1,o_t)$ is enforced to learn a compact representation from the given observations, $o_1$ and $o_T$, to convey the desired task, which serves as the contextual information.
\subsection{Generation Model}
\label{subsec:generate}
After inferring the context variable, the remaining question is: how to model $p(a_{1:T},s_{1:T}|z_c)$ to solve the planning problem? 
We assume the underlying process in Fig.~\ref{fig:showcase} is a fully observable Goal-conditioned Markov Decision Process $(\mathcal{S},\mathcal{A},\mathcal{T},\mathcal{R},\mathcal{C})$, where $\mathcal{S,A}$ is the state and action space.
We denote $p(a_t|s_t) $ as policy $\pi_\theta$ and $p(s_{t}|z_c,s_{t-1},a_{t-1})$ as transition model $\mathcal{T}_\mu$.
In this way, the generative model $p(a_{\sss{1:T}},s_{\sss{1:T}}|z_c)$ can be factorized as:
 \vspace{-2mm}
\begin{equation}
 \label{equ:MDP}
 \hspace*{-0.1cm} p(a_{\sss{1:T}},s_{\sss{1:T}}|z_c)=\prod^T_{t=1}\pi_\theta(a_t|s_t)\mathcal{T}_\mu(s_t|z_c,s_{t-1},a_{t-1})\;,
\end{equation}
where we use the convention that $s_0,a_0 = 0$.

A popular way to solve the MDP problem is using RL algorithms.
However, we only have access to expert trajectories $\{(o_{\sss{1:T}}^j,a_{\sss{1:T}}^j)\}_{j=0}^K{\sim}\pi_{E}$ without a well-defined reward function, making it infeasible to directly apply RL algorithms.
Therefore, we adopt an IL approach and use the expert trajectories as demonstrations.
However, there are still several key difficulties:
\begin{enumerate*}[label=\alph*)]
    \item typical IL algorithm is model-free algorithm that is ideal for learning \textit{habitual} behavior without thought for actions' consequences~\cite{liljeholm2018instrumental}, making it imperfect to learn goal-directed actions.
	\item the static dataset cannot provide feedback signals or the transferred states as the learning agent interacts with it; and
	\item each demonstration trajectory is  performed by the expert to reach a specific goal and thus might not be sufficiently explored under different situations.
\end{enumerate*}
Below, we address these difficulties.

\noindent \textbf{Effective Imitating with Transition Model.}
Instead of short-term environment learning as in \cite{chang2019procedure}, we optimize model with the whole trajectory. 
Inspired by GAIL~\cite{NIPS2016_6391}, we formulate the IL problem as an occupancy measure matching problem~\cite{ho2016model}, where the goal is to minimize the Jenson-Shanon divergence of trajectory distributions induced by the learned policy $\pi_\theta$ and the expert policy $\pi_E$ respectively.
In order to learn goal-directed actions, and bypass the need of an interactive environment which is required the original GAIL, we employ a transition model to roll out and jointly optimize it with policy learning.
There are two important reasons for the joint optimization:
\begin{enumerate*}[label=\alph*)]
	\item during the training, the action policy is not stationary, which means a pre-trained transition model will not help the action policy explore better decisions; and 
	\item the transition model can interact with action policy so that the learned latent space is optimized on the entire state-action pairs induced by the expert policy $\pi_E$, which helps it incorporate information over multiple time steps.
\end{enumerate*}
The model can be either deterministic or stochastic; thus, we introduce two versions of the transition model.

\noindent \textbf{Int-MGAIL:} In Interior-Model GAIL, the transition model is built inside the LSTM cell, which can be seen as a fully deterministic model.
We modify the LSTM cell and treat the long-term cell state as the state $s_t$ in Eq.~\ref{equ:MDP} and the short-term hidden state as our action $a_t$, so that we can enforce the action 
\begin{enumerate*}[label=\alph*)]
\item to interact with the hidden state to roll out the next state
\item only depends on the current state.
\end{enumerate*}
At each time-step, the input to the cell is the previous cell's long-term and short-term state $s_{t-1}$ and $a_{t-1}$, as shown in Eq.~\ref{equ:cell}. 
  \begin{equation}
  	\label{equ:cell}
  	\begin{split}
  		f_t &=\sigma (W_f * a_t + U_f * s_t + b_f)\;,\\
  		i_t &=\sigma (W_i * a_t + U_i * s_t + b_i)\;,\\
  		a_t &= \textup{Tanh} (W_a *  s_t + b_a)\;,\\
  		s_{t+1}&=f_t*[s_t,z_c]+i_t *a_t \;.
  	\end{split}
  \end{equation}

\noindent \textbf{Ext-MGAIL:} Int-MGAIL provided a deterministic solution for modeling the unknown environment, but it will underestimate the uncertainty of the environment. 
Therefore, we further take the transition model as an external module to explicitly model the environment transition in a stochastic way, meaning different observations can follow the same state.
To model the uncertainty, we designed the action policy as a stochastic model with Bernoulli probability vector of $a_t$ because of the discrete action space. 
The stochastic modeling is crucial for successful planning when we have the same start and goal states but different procedures.

\noindent \textbf{Hindsight Relabeling.}
The problem with the static dataset is that each episode only shows one possible way to reach the specified goal, which limits the agent's ability to explore what would happen had the circumstance been different.
Inspired by HER~\cite{andrychowicz2017hindsight}, we utilized the relabeling method that tries to alleviate this problem by augmenting the demonstrations with ``fake'' goals that were attained in the episode, allowing the agent to sufficiently explore the state-action space and make better decisions for the future.
Formally, we have one valid trajectory $\{(o_{\sss{1:T}}^j,a_{\sss{1:T}}^j)\}$ of an expert attempting to reach the goal $o_{\sss T}$ at $j$th episode from the start.
Then, the portion of this trajectory between any two non-adjacent observations, $o_m$ and $o_n$, can also be seen as a valid trajectory as the expert attempts to reach $o_n$ starting from $o_m$.
Therefore, for every trajectory in the original dataset, we select two non-adjacent observations and augment the dataset with $D\leftarrow D\cup{(o_{m:n},a_{m:n})}$.
The intuition behind this process is that we can replay each episode with a different goal than the one the expert was initially trying to achieve.
\subsection{Learning}
We have three main components to be optimized:
\begin{enumerate*}[label=\alph*)]
	\item the transition model  $\mathcal{T}_\mu(s_{t+1}|s_t,a_t,z_c)$ that uses previous state-action pair and context variable to predict the next state;
	\item the policy model $\pi_\theta(a_t|s_t)$ that models the distribution over the set of action under current state; and 
	\item the discriminator $\discriminator$, parameterized by $\omega$, tries to distinguish the $\{(s_t,a_t)\}$ from the expert or the learned policy $\pi_\theta$.
\end{enumerate*}

We refer the expert trajectory as $\tau^E= \{(s^{\sss E}_t,a^{\sss E}_t)\}$ and trajectory $\tau= \{(s_t,a_t)\}$ as state-action pairs visited by the current learned policy.
We first randomly sample $\tau^E$ from the dataset and roll out  $\tau^E$ accordingly, then we optimize the discriminator by ascending the gradient in Eq.~\ref{loss_D}:
\begin{equation}
	\label{loss_D}
	\mathbb{E}_{\pi_\theta} [\nabla_\omega\log(1-\discriminator(s_t,a_t))] +
	\mathbb{E}_{\pi_{\sss E}} [\nabla_\omega\log(\discriminator(s^{\sss E}_t,a^{\sss E}_t))]\;.
\end{equation}
We further let the discriminator gradient back-propagate into the previous time-step, helping the transition model to learn the further consequences related to current action.
However, we observe a high variance problem during the training.
Therefore, we employ an additional loss to help the generated states $s_t$ quickly move to regions close to the expert-visited states.
Hence, the transition model is optimized by descending the gradient in Eq.~\ref{loss_T}:
\begin{equation}
	\label{loss_T}
	\begin{split}
		&\mathbb{E}_{a_t\sim \policy}[\nabla_\mu\log(1-\discriminator(\transition(s_{t-1},a_{t-1},z_c),a_t^{\sss E}))] \\
			&+\mathbb{E}_{s_t^{\sss E}\sim\pi^{\sss E}}[\nabla_\mu \mathcal{L}(\transition(s_{t-1},a_{t-1},z_c),s_t^{\sss E})]\;,
				\end{split}
\end{equation}
where $\mathcal{L}$ measures the distance between two latent vectors.
								
The last component is the action policy $\pi_\theta$.
After optimizing the discriminator, we can interpreted it as a local reward function and we optimize policy to maximum the reward $r(s_t, a_t) = log(D_\omega(s_j, a_j))$.
In order to imitate the expert rather than mimicking, the action policy needs the ability to intentionally explore actions that the expert did not perform.
We adopted the offline policy evaluation in offline-RL and follow the classical evaluation method~\cite{hanna2019importance,degris2012off}, re-weighting the rewards by the importance sampling ratio (Eq.~\ref{loss_P}) to select a better policy during training. 
Concretely, we first learn a classification network as the behavior policy $\beta(a_t|s_t)$ from demonstrations via behavioral cloning.
Then optimize $\pi_\theta$ with policy gradient which tries to maximize the accumulated reward along the whole trajectory:
\begin{equation}
	\label{loss_P}
	\mathbb{E}_{\beta} [\frac{\policy(a_t|s_t)}{\beta(a_t|s_t)}\nabla_\theta\log\policy(a_t|s_t) Q(s_t,a_t)]-\lambda \mathcal{H}(\policy)\;,
\end{equation}
where $\mathcal{H}(\policy)=\mathbb{E}_{\policy}[-\log\policy(a|s)]$ is the policy entropy.

\begin{algorithm}[t]
	\hspace*{0.02in} {\bf Input:}
	All observations $\{o_i\}^N_{i=1}$, set of action $\{a_i\}_{i=1}^M$, models $\transition , \policy$ planning length $T$
	\begin{algorithmic}[1]
		\caption{Walk-through Planning}
		\label{alg:walk}
		\State Initialize observation list  $\beta \gets \emptyset$
		\For{$i = 1, 2,\cdots, N$}
		\State $s_i=\phi(o_i)$
		\EndFor
		\For{$i = 1, 2,\cdots, N$}
		\State $s_{\rm next}= \transition(s_i,\policy(s_i))$
		\LineComment{Find the index of the nearest state}
		\State $ k=\underset{k}{\arg\min} \lVert s_{k} - s_{\rm next} \rVert_2^2 $
		\LineComment{Increment the transition probability over all action}
		\State $ S_{i,k}\mathrel{+}=\sum\limits_{m=0}^M \policy(a_m|s_i)$
														
		\EndFor
		\State $\beta \gets \underset{\rho \in Perm(T)}{\arg\max} \sum\limits_{i=1}^T S_{i,\rho(i)}$
	\end{algorithmic}
\end{algorithm}

\subsection{Walk-through Planning with Transition Model}
\label{sec:method:planning}
Given the start and goal observations, we first infer the contextual information by sampling from $\posterior(z_c|o_1,o_T)$.
Based on the sampled $z_c$, the generation model will roll out subsequent actions and hidden states as the sampled trajectory.
Given the pool of visual observations $\{o_i\}$, we first construct the score matrix $S_{i,j}$ to capture the transition probability between $o_i$ and $o_j$ with the sampled trajectory, as shown in Alg.~\ref{alg:walk}.
After constructing the rank score table, we can then perform walk-through planning to retrieve the intermediate observations that lead to the goal.
As suggested in~\cite{chang2019procedure}, this problem can be seen as ﬁnding a permutation function $b$ : $\{ 1, 2, \cdots, T \} \rightarrow \{ 1, 2, ..., T \}$ that maximizes the transition probability along the permutation path, subject to the constraints that $b(1) = 1, b(T) = T$.

\section{Experiments}
\label{sec:exp}
\begin{figure}[t]
   \centering
	\includegraphics[width=0.45\textwidth]{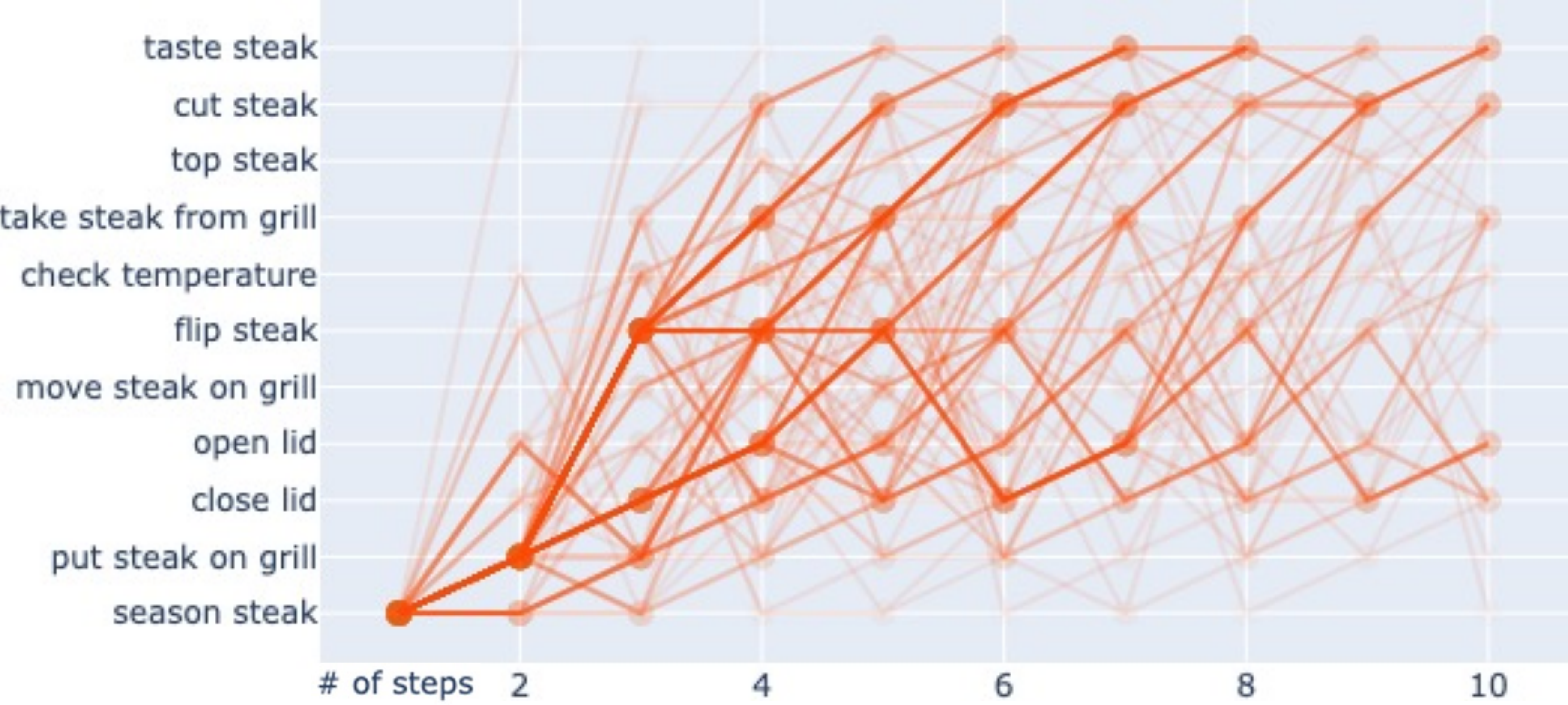}
	\caption{\textbf{Expert trajectories of \textit{Grill Steak} task.} Heavier color indicates more frequently visited path}
	\label{fig:stat}
	\vspace{-6mm}
\end{figure}

We choose CrossTask~\cite{crosstask} to conduct our experiments, which consists of 2,750 video (212 hours in total). Each video depicts one of the 18 primary long-horizon tasks such as \textit{Grill Steak} or \textit{Make French Strawberry Cake}.
For the videos in each task, we randomly select 70\% for training and 30\% for testing.
Different tasks have various procedure steps: less complex tasks include \textit{jack up a car} (3 steps); more complex ones include \textit{pickle cucumbers or change tire} (11 steps), and the steps do not necessarily appear in the same order as the task description as shown in Fig.~\ref{fig:stat}.

Each video has densely annotated boundaries with caption labels that describe the person's actions in the video.
We treat each video as a sequence of images $I_{1:N}$ having annotated description $v_{1:M}$ with temporal boundaries $(s_{1:M},e_{1:M})$.
For $i$-th video clip, we choose frames around the beginning of the captions $I_{s_i - \delta:s_i+\delta}$ as $o_i$, caption description $v_i$ as the semantic meaning of action, and images nearby the end  $I_{e_i-\delta:e_i+\delta}$ as the next observation $o_{i+1}$.
Here, $\delta$ controls the duration of each observation, and we set $\delta=1$ for all experiments.
We further use the relabeling technique introduced in Section~\ref{subsec:generate} to augment the data with randomly selected 30\% of the expert trajectories.
								
To construct our state-space $\mathcal{S}$, we use pre-computed features provided in CrossTask as our state estimation: One second of the video is encoded into a 3,200-dimensional feature vector which is a concatenation of the I3D, Resnet-152, and audio VGG features~\cite{hershey2017cnn,he2016deep,carreira2017quo}.
Note here we do not use the state estimations for testing; we only use them for training the Generation model.
Lastly, we construct the action space $\mathcal{A}$ by enumerating all combinations of the caption description's predicates and objects, which provides 105 action labels and are shared across all 18 tasks.

\noindent \textbf{Implementation Details.}
For computing context variables, we use the DCGAN architecture~\cite{radford2015unsupervised} as the image encoder and decoder in our model.
The behavior policy is a classification network that takes state estimation as input and generates the probability over the action space.
The policy network for both Int-MGAIL and Ext-MGAIL share a similar structure as the off-policy Actor-Critic network \cite{degris2012off}, which is two-headed: one for computing an action based on a state and another one producing the expected return values of the action.
In the Ext-MGAIL, we assume our hidden state $s_t$ to be Gaussian; thus, the transition model is Gaussian with mean and variance parameterized by a feed-forward neural network.
The discriminator networks for both models share the same architecture, which is a similar network in the original GAIL~\cite{NIPS2016_6391}.
Further implementation details can be found in the supplementary material.
\subsection{Evaluating Procedure Planning}
\label{sec:exp:pp}
We compared with the following methods:

\noindent \textbf{- Uniform Policy.} At each step, the algorithm will uniformly sample one action from all actions. This method serves as the empirical lower bound of performance.
									
\noindent \textbf{- Universal Planning Networks (UPN)}~\cite{srinivas2018universal}. Like our method, UPN learned a plannable latent representation where gradient descent can be used to compute a plan that minimizes a supervised imitation loss. We expand the original UPN to discrete action space by using a softmax layer to output probability over discrete actions.
									
\noindent \textbf{- Dual Dynamics Networks (DDN)}~\cite{chang2019procedure}. DDN is the first work proposing procedure planning in instructional video problem. Similar to UPN, it learns the dual dynamic of the state-action transition and perform sample-based planning upon the learned latent representation.

When evaluating with the pre-collected dataset, a common way is the re-weighted rewards~\cite{levine2020offline}.
But there is no defined rewards here. 
To keep consistent with state-of-the-art methods, we use three different matrices to evaluate the performance and limit the experiment to length 3-5 even our method is applicable for longer trajectory modeling.

\noindent  \textbf{- Success Rate} is designed to evaluate the long-term action association, \ie, the correctness of \textit{action sequence}. Only if every action matches, this plan is considered as a success. The result is the percentage of the successes in test dataset.
									
\noindent \textbf{- Accuracy} is used to evaluate the correctness of \textit{individual time step action}, which serves as a constraint relaxation of the success rate metric. The individual action is considered as a success only if it matches the ground truth at the same time step, which is written in terms of percentages.
									
\noindent \textbf{- mIoU} is used to capture the cases where the model can output the right actions but \textit{fail to preserve the actions' order}. We adopt this metric from \cite{chang2019procedure} that computes IoU $\frac{|\{a_t\}\cap \{\hat{a_t}\}|}{|\{a_t\}\cup \{\hat{a_t}\}|}$ between the set of ground-truth $\{ a_t \}$ and the planned actions $\{\hat{a_t}\}$.

As illustrated in Table~\ref{table:PP}, UPN can learn representations that perform reasonably well compared to the uniform baseline.
However, as instructional videos' action space is not continuous, the gradient-based planner cannot work well.
The proposed Int-MGAIL outperforms baseline DDN at two different time-scales. The reason is that we perform RL training, which maximizes the accumulated reward alone the whole trajectory. 
By introducing the stochastic process into the action policy, our Ext-MGAIL has a better performance.
This is because, given the same beginning and goal observation, there is more than one valid sequence of actions.
By designing a model with both stochastic and deterministic components, we show that our agent successfully learns plannable representations from real-world videos to outperform all the baseline approaches on all metrics.

\begin{table}[t]
	\caption{\textbf{Results of Procedure Planning.} Our models signiﬁcantly outperform the baselines by $\sim10\%$ improvement in terms of the success rate. Our Ext-MGAIL has a marginal improvement compared with Int-MGAIL; this shows that introducing a stochastic process in the policy can help the policy explore and thus improve the performance.}
	\begin{tabular*}{\linewidth}{@{\extracolsep{\fill}}rl|ccccc}
		\toprule
		&~ & Uniform & UPN & DDN & Int & Ext \\ \midrule
		&Succ. & 0.01 & 2.89 & 12.18 & 17.03 & \textbf{21.27} \\
		${\scriptstyle {\rm T}=3}$&Acc. & 0.94 & 24.39 & 31.29 & 44.66 & \textbf{49.46} \\
		&mIoU & 1.66 & 31.56 & 47.48 & 58.08 & \textbf{61.70} \\ \midrule
		&Succ.& 0.01 & 1.19 & 5.97 & 9.47 & \textbf{16.41} \\
		${\scriptstyle {\rm T}=4}$&Acc & 0.83 & 21.59 & 27.10 & 37.16 & \textbf{43.05} \\
		&mIoU & 1.66 & 28.85 & 48.46 & 57.24 & \textbf{60.93}\\
		\bottomrule
	\end{tabular*}
	\label{table:PP}
	\vspace{-5mm}
\end{table}
		
\begin{figure}[h]
	\includegraphics[width=\columnwidth]{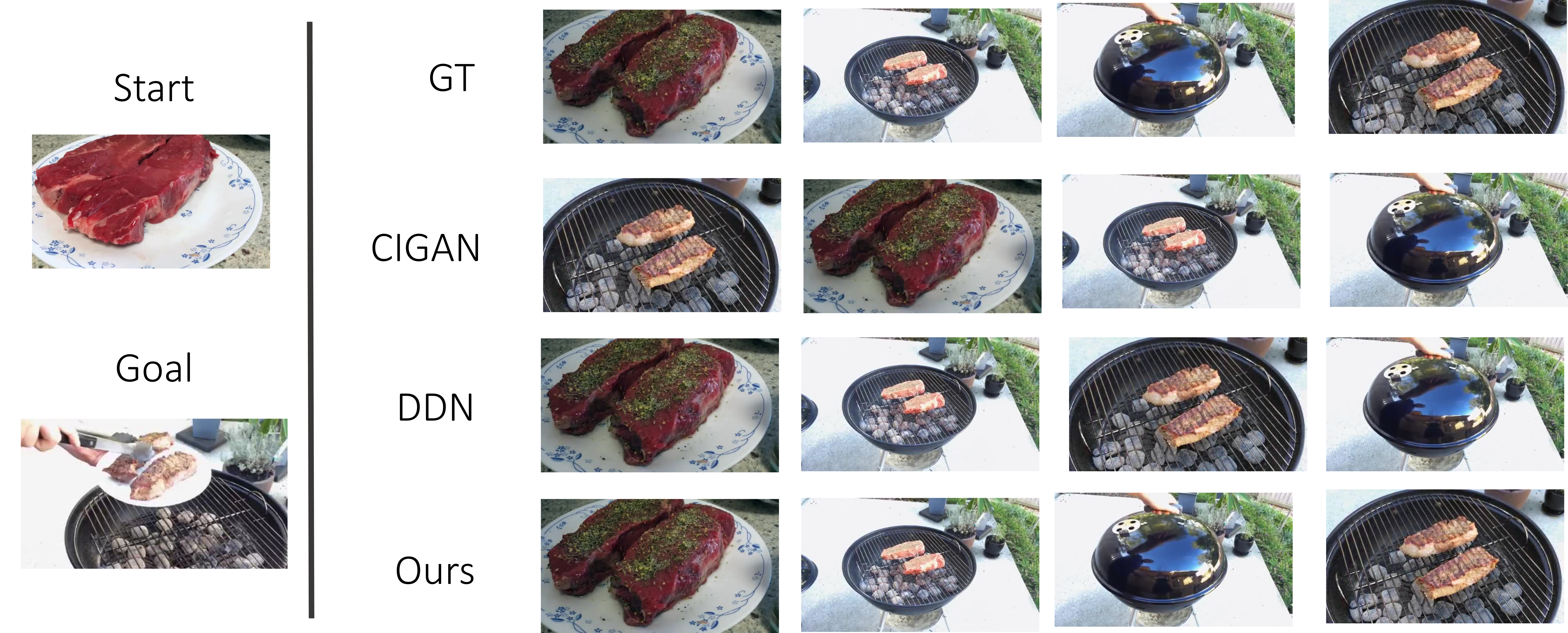}
	\caption{\textbf{Walk-through Planning qualitative results for \textit{Grill Steak}}. Given the starting and goal observations, our model can output the correct order for each step.}
	\label{fig:walk}
	\vspace{-5mm}
\end{figure}
		
\subsection{Evaluating Walk-through Planning}
\label{sec:exp:wp}
								
Different from procedure planning, given the the observations $o_1$  $o_{\sss T}$, the model needs to generate the intermediate observations $\{o_2,\cdots,o_T\}$.
Directly predicting the raw RGB image is unnecessarily hard~\cite{finn2016unsupervised}.
Similar to the setup in~\cite{chang2019procedure}, instead of generating frames, we retrieve the intermediate video clips from dataset in the correct order.
To find the path from $o_1$ to $o_{\sss T}$, a rank table is constructed to evaluate the transition probability between two clips.

In addition to the \textbf{Uniform} policy and \textbf{DDN}, we include the \textbf{Causal InfoGAN (CIGAN)} into comparison. 
Like our approach, they plan the trajectory in latent space but use the generative model to transform the trajectory to observations directly. 
The advantage of CIGAN is that it can be trained to perform walk-through planning without action supervision.

The evaluations are conducted on the following metrics.
\noindent \textbf{- Hamming.} As described earlier, we are finding the best permutation of the observation index. Then the distance is defined as $d(y,\hat{y})=\sum_{i=1}^TI(1|y_i\neq\hat{y}_i)$, which is good for evaluating the \textit{single step observation order}.
				
\noindent \textbf{- Pair Accuracy.}	To compare the distance between two permutation sequence, we use pairwise accuracy to calculate the distance along the planned and ground truth observation orders. 
This is defined as $\frac{2}{T(T-1)}\sum_{i<j,i\neq j}^T I(1|y_i<\hat{y}_j)$.
								
The results are shown in Table \ref{table:WP} and Fig.~\ref{fig:walk}. 
CIGAN can learn reasonable models beyond Uniform without using action supervision.
However, the complexity of the instructional videos requires explicit modeling of the forward dynamics conditioned on the semantic actions.
Our two methods outperform all baseline models, which means both models are applicable to both planning and walk-through planning.
We also show that effective IL requires learning a transition model and optimizing policy on multi-step transitions instead of individual state-action pairs.
						
\begin{table}[t]
	\caption{\textbf{Results of Walk-through Planning.} Our model outperforms the baselines by explicitly modeling the transition dynamics between temporally adjacent observations.}
	\begin{tabular*}{\linewidth}{@{\extracolsep{\fill}}rl|ccccc}
		\toprule
		&~&Uniform & UPN & DDN & Int & Ext \\ \midrule
		\multirow{2}{*}{${\scriptstyle {\rm T}=3}$}&Ham.& 1.06 & 0.57 & 0.26 & 0.19 & \textbf{0.13} \\
		&Pacc. & 46.85 & 71.55 & 86.81 & 86.98 & \textbf{93.66} \\\midrule
		\multirow{2}{*}{${\scriptstyle {\rm T}=4}$}&Ham.& 1.36 & 1.36 & 0.88 & 0.70 & \textbf{0.57} \\
		&Pacc. & 52.23 &68.41 &  81.21 & 86.42 & \textbf{89.74}\\
		\bottomrule
	\end{tabular*}
	\label{table:WP}
\end{table}

\subsection{Visualization of Contextual Information}
\begin{figure}[t]
	\centering
	\includegraphics[width=0.45\columnwidth]{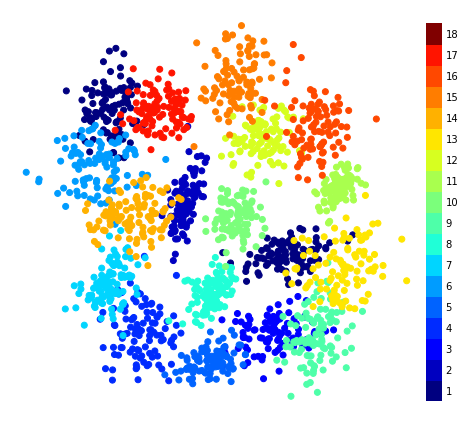}
	\caption{\textbf{Visualization of the contextual information} of the starting and goal observations of all 18 tasks in the CrossTask dataset. The colorbar indicates the ground-truth task labels, where different colors indicate different tasks.}
	\label{fig:latent}
	\vspace{-5mm}
\end{figure}

In this section, we aim to answer the following two questions:
\begin{enumerate*}[label=\alph*)]
	\item Can the proposed Inference model learn the useful contextual information of different tasks from the demonstrations?
	\item Why does the contextual information help the subsequent action learning?
\end{enumerate*}
To this end, we use t-SNE~\cite{maaten2008visualizing} to reduce the dimension of the context variable $z_c$ to 2 and visualize $z_c$ all 18 tasks as shown in Fig.~\ref{fig:latent}.
For every task, we randomly sampled 100 pairs of start and goal clips (1800 pairs in total) and extracted their contextual information.

As we can see, all the samples are grouped by the tasks' labels, which suggests the Inference model has learned roughly distinct regions in the hidden space to correspond to each task in the dataset.
Further task descriptions can be found in the supplementary material.
Considering we never used the task labels in the learning process, this result indicates an underlying relationship between the different tasks' observations.
This suggests that the Inference model can encode sufficient information to convey the desired task.
Furthermore, the Generation model can benefit from this concise embedding on modeling decision-making because irrelevant deviations from the raw pixel space will create exponentially diverging trajectories.

However, there still exist overlaps between different clusters, and some samples drift from the majority.
This reflects the fact that we cannot entirely rely on the contextual information to recover the expert's decision process, which further validates the effectiveness of the Generation model.

\subsection{Ablation Study}
We conducted experiments with three variations:
\noindent \textbf{w/o r:} The model is trained with both sequence mapping loss and discriminator loss but w/o maximizing accumulated reward. In this way, the model tries to match the short-term actions w/o considering the trajectory as whole.
\noindent \textbf{w/o dis:} We further drops the discriminator loss in Eq.~\ref{loss_T}, making it a supervised seq2seq learning model.
As shown in Table \ref{table:AS}, learning the model solely w/o dis will significantly hurt the overall performance because the model simply optimized on expert trajectory will be over-fitted to the regions that the expert traversed and thus make it hard to generalize to other areas that might be helpful for action policy learning.
We think the main reason for the low success rates with sequence mapping is that pure supervised learning will excessively focus on recovering the same action sequence as ground-truth, ignoring the fact that actions can be exchanged to achieve the same goal.
\noindent \textbf{w/o HER:} We observed that combining HER always brings performance boosts. Our stochastic $\textrm{Ext}_{\textrm{w/o HER}}$ has a lower performance than Int-MGAIL; we suspect that it is because, without HER, the original dataset is insufficient for stochastic model optimizing the whole trajectory. This result is consistent with our observation on the experiment with the additional dataset.
\begin{table}[h]
	\centering
	\scalebox{0.8}{
		\begin{tabular*}{1.2\linewidth}{@{\extracolsep{\fill}}rl|cccccc}
			\toprule
			& ${\scriptstyle {\rm T}=3}$ 		& 	\thead{Int \\w/o r} 	& \thead{Int \\w/o dis}	& \thead{Ext \\w/o r}   & \thead{Ext \\w/o dis}    &\thead{Int \\w/o HER} & \thead{Ext \\w/o HER}           \\ \midrule
			&   Succ. 	&    7.18 					&  5.89 				& 15.18 				& 11.42                &         14.39      &  18.01  \\
			&   Acc. 	&   18.74 					& 11.66 				& 27.29 				& 23.46    &  37.43  &  43.86  \\
			&   mIoU 	&   27.51 					& 20.66 				& 37.48 				& 30.97                  &  54.18             &   57.16      \\
																	
			\bottomrule
		\end{tabular*}}
	\caption{\textbf{Results of Ablation Study.}  The performance decreases signiﬁcantly when optimizing without sequence modeling or only with a supervised learning loss, which shows the importance of learning over the whole trajectory.}
	\vspace{-5mm}
	\label{table:AS}
\end{table}
\subsection{Addtional Experiment}
We further compare our model with previous best performing method DDN on a second dataset~\cite{alayrac2016unsupervised}.
Both of our models outperformed the DDN, but the increase of the accuracy on procedure planning (Succ.${\sim}4$\%, T=3) is smaller than one obtained on the CrossTask, and the performance between Int-MGAIL and Ext-MGAIL are very similar (Succ. 20.19/22.11\%, T=3). 
We suspect that it is because the new dataset does not provide sufficient samples for optimizing long trajectories.
More details can be found in the supplementary material.

\section{Conclusion, Application, and Future Work}
In this paper, we present a new method to address the procedure planning problem focusing on learning goal-directed actions.
Concretely, we propose a predictive VAE structure that learns to embed the contextual information of the desired task. 
Moreover, we propose two novel model-based imitation-learning algorithms to solve the formulated decision-making problem in unknown environments.
Results on real-world instructional videos show that our approach can learn a meaningful representation for planning and uncover the human decision-making process.

Being able to learn goal-directed actions from the pixels, the proposed method enables the AI system to extract useful information from expert demonstrations.
Moreover, learning policy from the offline dataset avoids online interaction with the environment, making our method practical in real-world applications, \eg, service robots.

A direction of future work is investigating different ways to combine contextual information with environment dynamics, such as through self-supervision.
Another important future direction is to consider the collected data as policy constraints such that robots can act safely in the real world and continuously improve themselves by accumulating data of the environment interactions, making robotic agents more capable of solving challenging real-life tasks.

\vspace{2mm}
\noindent \textbf{Acknowledgments.} This work has been partially supported by the National Science Foundation (NSF) under Grant 1813709 and the National Institute of Standards and Technology (NIST) under Grant 60NANB17D191. The article solely reflects the opinions and conclusions of its authors but not the funding agents.

{\small
	\bibliographystyle{ieee}
	\bibliography{egbib}
}
				
\end{document}